\documentclass{article}

\usepackage{PRIMEarxiv}

\usepackage[utf8]{inputenc} 
\usepackage[T1]{fontenc}    
\usepackage{hyperref}       
\usepackage{url}            
\usepackage{booktabs}       
\usepackage{amsfonts}       
\usepackage{nicefrac}       
\usepackage{microtype}      
\usepackage{fancyhdr}       
\usepackage{graphicx}       
\graphicspath{{media/}}     
\usepackage{multicol}  
\usepackage{multirow}
\usepackage{soul}
\usepackage{subfigure}
\usepackage{graphicx}
\usepackage{textcomp}
\usepackage{xcolor}

\usepackage{algorithm}
\usepackage{algorithmic}

\title{POPNASv2: An Efficient Multi-Objective Neural Architecture Search Technique
}

\author{
  Andrea Falanti, Eugenio Lomurno, Stefano Samele,  \\
  Politecnico di Milano \\
  Milan, Italy\\
  \texttt{\{andrea.falanti, eugenio.lomurno, stefano.samele\}@polimi.it} \\
   \And
  Danilo Ardagna, Matteo Matteucci \\
  Politecnico di Milano \\
  Milan, Italy\\
  \texttt{\{danilo.ardagna, matteo matteucci\}@polimi.it} \\
}

\begin{document}
\maketitle

\begin{abstract}
Automating the research for the best neural network model is a task that has gained more and more relevance in the last few years. In this context, Neural Architecture Search (NAS) represents the most effective technique whose results rival
the state of the art hand-crafted architectures.
However, this approach requires a lot of computational capabilities as well as research time, which make prohibitive its usage in many real-world scenarios.
With its sequential model-based optimization strategy, Progressive Neural Architecture Search (PNAS) represents a possible step forward to face this resources issue. Despite the quality of the found network architectures, this technique is still limited in research time.
A significant step in this direction has been done by Pareto-Optimal Progressive Neural Architecture Search (POPNAS), which expand PNAS with a time predictor to enable a trade-off between search time and accuracy, considering a multi-objective optimization problem.\\
This paper proposes a new version of the Pareto-Optimal Progressive Neural Architecture Search, called POPNASv2.
Our approach enhances its first version and improves its performance. 
We expanded the search space by adding new operators and improved the quality of both predictors to build more accurate Pareto fronts.
Moreover, we introduced cell equivalence checks and enriched the search strategy with an adaptive greedy exploration step.
Our efforts allow POPNASv2 to achieve PNAS-like performance with an average 4x factor search time speed-up.
\end{abstract}


\section{Introduction}
Nowadays, deep neural networks represent the state of the art in solving complex tasks, like image classification, object recognition and text processing.
Their strength relies on automatically extracting patterns and rules from enormous data amounts thanks to the GPU-based computation acceleration.
Usually, deep neural networks are hand-crafted following intuition and empirical rules that lead to performance boosts in similar tasks.
However, despite design skills, human efforts are not always translated into good models.
In fact, searching an effective architecture for a specific domain is a challenging task, which can be very time consuming and not consistently achieves the desired performance requirements.

Many of the latest researches are focused on the automation of neural architecture engineering, i.e., the process of algorithmically designing deep neural networks to optimize their performance on a given dataset.
Zoph \textit{et al.}~\cite{zoph2017neural} have been among the first to formalize this task under the name of Neural Architecture Search (NAS).
In their version, the algorithm relies on the presence of three main components, namely, the search space, the search strategy and the performance estimation strategy. The goal is to build a cell, a basic unit constituted by different operations named blocks.
The successive repetition of these units creates the network architecture.

Despite the high quality of the neural networks obtained by exploiting the NAS technique, the enormous amount of time and computational resources required do not allow a widespread use of this approach.
Progressive Neural Architecture Search (PNAS)~\cite{liu2018progressive} came as an effective solution to this issue.
With its sequential model-based optimization search strategy, this algorithm iteratively increases the researched neural networks complexity.
The progressive exploration starts from the simplest cells of the search space, and then progressively expands them with new blocks, based on their quality estimated by an auxiliary model called predictor.

Pareto-Optimal Progressive Neural Architecture Search (POPNAS) \cite{10.1145/3449726.3463146} introduces a new time regressor to further improve the search speed. Between cells with the same accuracy, those generating slow networks are removed from the exploration step. POPNAS doubles the speed of the search at the cost of a 10\% reduced accuracy.

This paper presents a new version of Pareto-optimal Progressive Neural Architecture Search (POPNASv2). Our method solves the tradeoff introduced by POPNAS, reducing the total search and training time without any loss in accuracy. POPNASv2 best models rival the ones found by PNAS, but they are produced with almost 4 times less the amount of GPU hours.

\section{Related Works}
Recently, Neural Architecture Search has become a widespread topic in different data science fields~\cite{ren2020comprehensive,elsken2019neural,liu2021survey,ren2021comprehensive}.
Many works have defined the basis and the methodology of neural architecture search, most of them starting from the work of Zoph \textit{et al.}~\cite{zoph2017neural} based on reinforcement learning. 
Further works refined the methodology and explored other solutions concerning the search strategy: DARTS~\cite{liu2019darts} exploited gradient-based search, AmoebaNet~\cite{real2019regularized} instead was based on evolutionary algorithms, while NASNet~\cite{zoph2018learning} enhanced the reinforcement learning strategy.

DARTS inspired a new branch of works using one-shot models~\cite{pmlr-v80-bender18a}, also referred to as supernets. These are huge single model networks that embed all possible architectures in the search space as individual paths of it. 
One-shot models enable the exploration of the search space very efficiently through weight sharing. 
The performance of any network selected by the search strategy can be evaluated on the weights of the supernet, pruning the other paths not involved in the architecture. 
However, this methodology introduces new challenges since training these huge networks composed of multiple paths and layers, easily introduces weight co-adaptation and bad out-of-the-box accuracy without retraining the architectures. 
Examples of works using one-shot models are SMASH~\cite{brock2017smash} and ProxylessNAS~\cite{cai2019proxylessnas}.

Progressive Neural Architecture Search (PNAS)~\cite{liu2018progressive} is a more efficient version of NASNet~\cite{zoph2018learning} algorithm, based on sequential model-based optimization (SMBO)~\cite{10.1007/978-3-642-25566-3_40}. 
PNAS defines its cells as Direct Acyclic Graphs (DAGs) composed of blocks. 
Blocks are micro level units composed of two operations selected from a list, and a reduction function. 
The algorithm defines two types of cells: normal cells and reduction cells. 
Normal cells keep the same dimensionality between inputs and outputs, while reduction cells halve the spatial dimension and double the output depth compared to the input. 
The main difference between PNAS and NASNet is that the first algorithm searches a single cell structure instead of two distinct ones for normal and reduction cells.
This change did not affect the accuracy results but drastically improved the search efficiency. 

The progressive exploration starts from the simplest models of the search space, i.e., the cells containing only a single block.
The models are progressively expanded with new blocks, based on their quality, which is estimated by a surrogate model, referred to as predictor.
The predictor is trained on the results obtained by the child networks, and it is used to estimate the accuracy that an expanded cell will reach after training.
The predictor chosen by PNAS is an ensemble of 5 LSTM~\cite{hochreiter1997long} models.
The networks contained in each expansion are limited, and the process repeats cyclically until the cells are expanded to a target amount of blocks.

Pareto-Optimal Progressive Neural Architecture Search (POPNAS)~\cite{10.1145/3449726.3463146} expanded PNAS by addressing the search as a multi-objective optimization problem, trying to achieve a trade-off between training time and accuracy.
POPNAS operations are ordered and weighted based on the required time to perform them. 
The information are fed to a new surrogate model to estimate the training time of the cell expansions. 
This time predictor is implemented as a linear regression with non-negative least square (NNLS) model.
The accuracy predictor is based on LSTM, similar to PNAS. 
Through the combined usage of the time and accuracy predictors, it is possible to estimate the objectives of the optimization problem and build a Pareto front through a domination rule. 
The Pareto front built identifies the top K cells to be trained in the next step.
Compared to PNAS, the search process efficiency increases by a 2x factor. However, the top networks are characterized by a noticeable negative accuracy gap.

\section{Methods}
In this section we introduce the POPNASv2 method by describing its three main components.
In particular, Section~\ref{s:search_space} describes the search space explored by the algorithm. Section~\ref{s:search_strategy} defines the main steps of the search strategy and how they are interconnected.
Section~\ref{s:perf_eval} provides more information about the implemented predictors, used to evaluate the quality of the cells.

\subsection{Search space} \label{s:search_space}
POPNASv2 extends the original search space defined for PNAS. 
The goal of the algorithm is to find the optimal cell structure for the given task, so that it can be stacked to compose the best network architecture.

Cells are composed of micro-level units called blocks.
Each block is specified as a 4 elements tuple $(i_{1}, o_{1}, i_{2}, o_{2})$, where $i_{1}$ is the input of the operator $o_{1}$ and $i_{2}$ is the input of the operator $o_{2}$. The outputs of the two operators are joined via an addition operation.
The search space defines the set of possible operators, as well as their inputs.

Since limiting the search to the Pareto front provides major speed-ups in the search process, as demonstrated by POPNAS, we adopt a more extensive set of operations. 

The chosen operator set $\mathcal{O}$ consists of the 12 operators:

\begin{multicols}{2}
	\begin{itemize}
		\item 3x3 depthwise sep conv
		\item 5x5 depthwise sep conv
		\item 7x7 depthwise sep conv
		\item 1x3-3x1 stacked conv
		\item 1x5-5x1 stacked conv
		\item 1x7-7x1 stacked conv
		\item identity
		\item 1x1 conv
		\item 3x3 conv
		\item 5x5 conv
		\item 2x2 maxpool
		\item 2x2 avgpool
	\end{itemize}
\end{multicols}

Inside a cell, each block is enumerated by the index value $1 \leq c \leq b$, where $b$ is the current number of blocks inside the cell.
We define with $\mathcal{I}_{c}$ the set of all the possible inputs of a generic block in position $c$.
For $b=1$, only the inputs values coming from the previous cell output are available. 
We name this cells distance as lookback. 
From a network configuration point of view, a lookback equal to -1 represents a sequential connection between two consecutive cells.
Instead, a lookback equal to -2 represents a skip-connection that jumps one intermediate cell.
We constraint the search space to have at most a lookback equal to -2.
The input set progressively expands with the amount of blocks $b$, since a block in position $c$ can use any block in previous positions as input.

The amount of unique blocks generated in each training step depends only on the cardinality of the inputs and operators sets.
Initially, for $ b = 1 $, there are 300 unique blocks.
The search space cardinality is the number of possible cells structures contained in the search space. 
In our configuration, the upper bound not considering cell equivalences is around $ 10^{15} $ architectures.
The main challenge of POPNASv2 is to search through this vast space efficiently, training neural network models which are pretty variegate in the results since it aims to find a Pareto front of the time-accuracy optimization problem.

During the search, all the cells relevant for the current training step are stored in an encoded form, from which the CNN can be built. 
When a cell is selected for training, the model generator builds the network from scratch, using the cell specification info.
We implement normal and reduction cells as defined in PNAS.
In POPNASv2 the network is a sequence of $M$ motifs composed by $N$ normal cells followed by a single reduction cell. 
The last motif has no reduction cell and is followed by GAP and Softmax layers to compute the final output.

\subsection{Search strategy} \label{s:search_strategy}
POPNASv2 search strategy is articulated in the following phases:
\begin{itemize}
	\item the initial thrust, which is the training of an empty cell.
	\item the training step, in which the selected cells are built and trained.
	\item the expansion step, which is aided by two predictors and estimates the quality of all possible expansions of the cells trained in the training step. A Pareto front is built from these estimates, reducing drastically the number of cells considered.
	\item the exploration step, which is a conditional step with the goal of exploring inputs and operators underused in the selected Pareto front.
\end{itemize}
Algorithm~\ref{alg:popnasv2_search} provides the pseudo-code of the search strategy.

\begin{algorithm}[t!]
	\caption{POPNASv2 search strategy}
	\label{alg:popnasv2_search}
	\begin{algorithmic}[1]
		\REQUIRE $B$ (max num blocks), $ E $ (epochs), $ K $ (beam size),\\ $ J $ (exploration beam size), $ T $ (time constraint),\\ \textit{CNN-hp} (networks hyperparameters), \textit{dataset}.
		\STATE $S_{0}$ = empty cell
		\STATE $\mathcal{A}_{0}$, $\mathcal{T}_{0}$ = train-cells($S_{0}$, CNN-hp, dataset) 
		\STATE $ S_{1} = B_{1} $ 
		\STATE  $\mathcal{A}_{1}$, $\mathcal{T}_{1}$ = train-cells($S_{1}$, CNN-hp, dataset)
		\STATE dynamic-reindex = initialize-reindex($\mathcal{A}_{0}$, $\mathcal{T}_{0}$, $\mathcal{A}_{1}$, $\mathcal{T}_{1}$) 
		\FOR{ b = 2 : B }
		    \STATE $\mathcal{P}_{acc}$ = fit($\mathcal{A}_{0 \to b-1}$, $S_{0 \to b-1}$) 
			\STATE $ \mathcal{F}_{b-1} = $extract-features$(S_{b-1}) $
			\STATE $ \mathcal{P}_{time} = $fit$(\mathcal{T}_{0 \to b-1}, \mathcal{F}_{0 \to b-1}) $
			\STATE $ S_{b}' = $expand-cells$(S_{b-1}) $
			\STATE $ \mathcal{A}_{b}' = $predict$(S_{b}', \mathcal{P}_{acc}) $
			\STATE $ \mathcal{T}_{b}' = $predict$(S_{b}', \mathcal{P}_{time}) $
			\STATE $ S_{b}'', \mathcal{A}_{b}'', \mathcal{T}_{b}'' = $apply-time-constraint$(S_{b}', \mathcal{A}_{b}', \mathcal{T}_{b}', T) $
			\STATE $ S_{b} = $build-pareto-front$(S_{b}'', \mathcal{A}_{b}'', \mathcal{T}_{b}'', K) $
			\STATE $ \tilde{\mathcal{O}}, \tilde{\mathcal{I}_{b}} = $build-exploration-sets$(S_{b}) $
			\IF{$ |\tilde{\mathcal{O}}| > 0 \lor |\tilde{\mathcal{I}_{b}}| > 0 $}
				\STATE $ S_{b, exp} = $build-epf$(S_{b}'', \mathcal{A}_{b}', \mathcal{T}_{b}', S_{b}, J) $
			\ELSE
				\STATE $ S_{b, exp} = \{\} $
			\ENDIF
			\STATE $ \mathcal{A}_{b}$, $\mathcal{T}_{b}$ = train-cells$(S_{b} \cup S_{b, exp}$, CNN-hp, dataset)
		\ENDFOR
	\end{algorithmic}
\end{algorithm}

POPNASv2 starts by training the empty cell (lines 1, 2 in Algorithm 1), a straightforward model composed only by a GAP layer followed by Softmax. 
This step is referred to as ``initial thrust" and is necessary to set up the features used by the predictors. 
In fact, the measured training time $ t_{0} $ and accuracy $ a_{0} $ reached by this network can be considered as a common bias of all neural networks trained. 
The difference between the considered cell results $ (t, a) $ and the bias $  (t_{0}, a_{0}) $ gives a better estimate of the impact of the inputs and operators used on both metrics.

After the initial thrust, all possible unique cells with a single block are trained on the target number of epochs $ E $. 
When the training is complete, POPNASv2 saves the training time $ t $ and the best validation accuracy $ a $ reached by each cell since these metrics are required for training the predictors (line 4). This concludes the training step of the cells with $ b = 1 $.
For all the following training steps ($ b > 1 $), the search process will repeat cyclically until the target $ b = B $ is reached.

Both time and accuracy predictors are trained on the data related to all networks trained during the search (lines 7, 8, 9). 
After the predictors have been fully trained, they are used to estimate the training time $ \hat{t} $ and the validation accuracy $ \hat{a} $ of all the possible expansions (line 10) of the cells trained in the previous step (lines 11, 12). The expansion of a cell adds any valid block into an existing cell. 

As a first step, if the parameter $ T $, a time constraint that can be optionally defined to restrict the search on less demanding networks, is provided, POPNASv2 discards all the expanded cells that have $ \hat{t} > T $ (line 13). 

POPNASv2 looks for the best cells in the so-called expansion step by identifying the Pareto front (line 14) of the time-accuracy optimization problem, using $ \hat{a} $ and $ \hat{t} $ estimated by the predictors.
We implemented a mechanism to identify equivalent models to maximize the run efficiency. This technique avoids training multiple cell encodings corresponding to the same neural network architecture. 

Two blocks are defined as equivalent if $ (i_{1}, o_{1}) $ and $ (i_{2}, o_{2}) $ are specular since the join operator (add) is commutative.
In this case, only one of the two blocks is kept.
Two cells are equivalent if the list of blocks contained in one cell is a permutation of the other cell blocks. 
The algorithm checks for cell equivalences when building the Pareto front, avoiding the insertion of equivalent cells.

After this preprocessing step, the Pareto front is built (line 14). 
The cell with the highest predicted accuracy $ \hat{a} $ is inserted as the Pareto front first element to initialize the process. 
Then a domination rule is applied to build the rest of the Pareto front.
In detail, for each accuracy score achieved by the expanded cells, we define as dominant the cell with the lowest training time.
The Pareto front is thus composed only of dominant cells and is limited in cardinality up to $K$ elements.

POPNASv2 introduces a new step in the search method, called \textit{exploration step}.
This step aims to train a small supplementary set of architectures that contains different characteristics from the ones prevalent in the Pareto front. 
The Pareto front heavily exploits the information about the metrics retrieved during the previous training steps. However, the expanded cells could potentially benefit more from operators and inputs that did not perform well in previous iterations. 
Since the complexity of the models progressively increases with the number of blocks, exploring these underused elements of the search space can lead to potentially different results with a minimal impact on the search efficiency.

The exploration step is performed right after the expansion step until the last but one iteration. 
Initially, POPNASv2 creates the set of inputs and operators to explore (line 15). 
It begins by computing the utilization percentages of all inputs and operators in the Pareto front.

An operator $ o \in \mathcal{O} $ is inserted in the operator exploration set $ \tilde{\mathcal{O}}$ if its frequency inside the Pareto front cells is less than $\frac{1}{5|\mathcal{O}|}$.
Similarly an input $ i \in \mathcal{I}_{c} $ is inserted in the input exploration set $ \tilde{\mathcal{I}_{c}} $ if its frequency inside the Pareto front cells is less than  $\frac{1}{5|\mathcal{I}_{c}|} $. 

After defining both $ \tilde{\mathcal{O}} $ and $ \tilde{\mathcal{I}_{c}} $, the \textit{Exploration Pareto Front} (EPF) is computed if at least one of these sets is not empty (line 17).
Each selected cell is associated with an exploration score, which adapts over time while the EPF is built. 
The reason for dynamically changing the score is to balance the exploration of both inputs and operators sets, also trying to encourage the usage of all the values of an exploration set, if $ \hat{t} $ and $ \hat{a} $ are good enough to fit the Pareto front.
Then, the algorithm counts the total usage in the EPF of exploration inputs, namely $ |i_{exp}| $, and exploration operators, namely $ |o_{exp}| $.
The frequencies of each exploration input and exploration operator is respectively referred as $ i_{perc} $ and $ o_{perc} $.

The rules that attributes exploration points to a cell are described as follows:
\begin{itemize}
	\item +1 for each input $ i \in \tilde{\mathcal{I}_{c}} $, with bonus:
	\begin{itemize}
	    \item +2 if $ i_{perc} \le \frac{1}{|\tilde{\mathcal{I}_{c}}|} $
	    \item +1 if $ |i_{exp}| \le |o_{exp}| $
	\end{itemize}
	\item +1 for each operator $ o \in \tilde{\mathcal{O}} $, with bonus:
	\begin{itemize}
	    \item +2 if $ o_{perc} \le \frac{1}{|\tilde{\mathcal{O}}|} $
	    \item +1 if $ |i_{exp}| \ge |o_{exp}| $
	\end{itemize}
\end{itemize}

If the score is $ > 0 $, further points can be assigned based on the difference between the predicted time and predicted accuracy of the considered cell and the ones of the last element of the EPF. In particular, one point is added for each 4\% of relative accuracy difference and for each 10\% of relative time difference.

To be accepted for the insertion in the EPF, a cell must have a score $ \ge $ 8 if both $ \tilde{\mathcal{O}} $ and $ \tilde{\mathcal{I}_{c}} $ sets are populated, $ \ge $ 4 instead if one of them is empty. Since initially the EPF is empty, the two additional rules are not used, but they are activated from the evaluation of the second element.

The total number of elements of the EPF is capped by the $ J $ parameter. When the exploration step ends, networks built using cells coming from both the standard Pareto front and the EPF are trained (line 21). The process repeats until the target amount of blocks $ B $ is reached.

\subsection{Performance estimation strategy} \label{s:perf_eval}
Given the enormous size of the search space, we argue that networks time and accuracy forecasts must be estimated as accurately as possible.
As in POPNAS, in POPNASv2 we have implemented two additional agents, referred to as \textit{accuracy predictor} and \textit{time predictor}, to estimate respectively cells accuracy $\hat{a}$ and time $\hat{t}$ with which populate the Pareto front.
At each step, both predictors are trained with the results of all the cells trained during all the previous searches. When the predictors are ready, the expansion step can start.

\subsubsection{Accuracy predictor}
The accuracy predictor is inspired by PNAS, which used an ensemble of 5 LSTM models. 
In our algorithm, the LSTM-based model receives two inputs, based on the cell encoding: a tensor with the blocks inputs, grouped per block, and a tensor with the blocks operators, also grouped per block. 
Both tensors have dimension $ (B, 2) $, since each block has 2 inputs and 2 operators and the maximum amount of blocks present in a cell is equal to $ B $, the target amount of blocks of the search run. 
Both input and operator values are encoded as 1-indexed categorical. If the considered cell has fewer blocks than $ B $, both tensors are padded with (0, 0) for each missing block.

The two tensors are then processed separately with an embedding layer and finally concatenated together. This new tensor is then processed by two different Conv1D layers, which produce the Q and K tensors of an attention layer that follows the convolutions.
The output of the Attention is finally used as input for a bidirectional LSTM, which produces the final hidden layer used by the sigmoid unit to predict the final value, the estimated accuracy $ \hat{a} $ of the cell. 
POPNASv2 creates 5 models with this structure, trains them via 5-fold cross-validation and finally builds up an ensemble by averaging models predictions.

\subsubsection{Time predictor}
As in POPNAS, the time predictor is the agent responsible for estimating the training time $ \hat{t} $ that a cell would require to train for $ E $ epochs. 

The features set used to train the time predictor is described below:
\begin{itemize}
	\item number of blocks
	\item number of cells
	\item the sum of the dynamic reindex value of each cell block operator (OP score)
	\item number of concatenated tensors in cell output
	\item usage of multiple lookbacks (boolean)
	\item cell DAG depth (in blocks)
	\item number of block dependencies
	\item \% of the total cell OP score related to the heaviest cell path
	\item \% of the total cell OP score related to the blocks using lookbacks as input.
\end{itemize}

The dynamic reindex is a metric introduced in POPNAS to evaluate and rank single operators impact with respect to their training time. 
In order to achieve this goal, a preliminary evaluation is carried out on cells composed of single blocks containing two identical operators.

In our work we implemented a revised version of dynamic reindex which takes into account time biases due to neural network components present in all the generated configurations.
In detail, the time $ t_{0} $ is due solely to GAP and Softmax layers and to the data augmentation process, which are common to all architectures, therefore excluding $ t_{0} $ provides a more fair estimation of the impact of the operators on the training time.
For each operator $ o \in \mathcal{O} $, considering $ t_{o} \in T $ the time taken to train the symmetric flat cell with encoding [(-1, o, -1, o)], the corresponding dynamic reindex value is computed as:
\begin{equation}
	index_{o} = \frac{t_{o} - t_{0}}{max(T) - t_{0}}
\end{equation}

In order to select the best time predictor, we tested the same models investigated in POPNAS, i.e., Non-Negative Least Square (NNLS), Ridge Regression and XGBoost, and we extended the analysis to CatBoost~\cite{prokhorenkova2019catboost}, considered among the most advanced gradient boosted techniques.

\begin{table}[t!]
	\caption{Comparison between the tested machine learning models candidate as time predictors. All the algorithms are evaluated over CIFAR10 dataset.}
	\centering
	\resizebox{\linewidth}{0.92\height}{%
	    \begin{tabular}{|c||cccc||cccc|}
    		\hline
    		\multirow{2}{*}{Model} & \multicolumn{4}{c||}{MAPE(\%)} & \multicolumn{4}{c|}{Spearman($ \rho $)} \\
    		& b=2 & b=3 & b=4 & b=5 & b=2 & b=3 & b=4 & b=5 \\
    		\hline
    		NNLS & 24.525 & 16.168 & 12.712 & 13.531 & 0.776 & 0.984 & 0.990 & 0.989 \\
    		Ridge & 26.852 & 23.648 & 7.793 & 6.326 & 0.827 & 0.993 & $ \mathbf{0.991} $ & 0.987 \\
    		XGBoost & 24.419 & 9.073 & 9.797 & 5.554 & 0.977 & $ \mathbf{0.996} $ & 0.953 & 0.986 \\
    		CatBoost & $ \mathbf{23.571} $ & $ \mathbf{6.298} $ & $ \mathbf{7.668} $ & $ \mathbf{2.852} $ & $ \mathbf{0.988} $ & $ \mathbf{0.996} $ & 0.99 & $ \mathbf{0.992} $ \\
    		\hline
	    \end{tabular}%
	}
	\label{tab:time_regressors_comparison}
\end{table}

Each of these machine learning models has been trained using random search for model hyperparameters tuning. This optimization step performs multiple trainings on 5-folds, using early stopping to preventively terminate the runs with bad hyperparameters, making the process more time-efficient.
We conducted all these ablation studies over CIFAR10 dataset~\cite{Krizhevsky09learningmultiple} which results are displayed in Tab.~\ref{tab:time_regressors_comparison}.

POPNASv2 relies on the estimations made by the predictors to build an effective Pareto front. The domination rule compares the estimated accuracy and time of two cells; therefore, it is essential to rank precisely the estimated values. We use Spearman's rank correlation coefficient ($ \rho $) to measure the quality of the ranking of a predictor.
The accuracy of the predictors (measured with Mean Absolute Percentage Error - MAPE) is less relevant than high-quality ranking. However, it is preferred to enable the time constraint $ T $.

\begin{figure}[t!]
	\centering
	\includegraphics[width=0.65\linewidth]{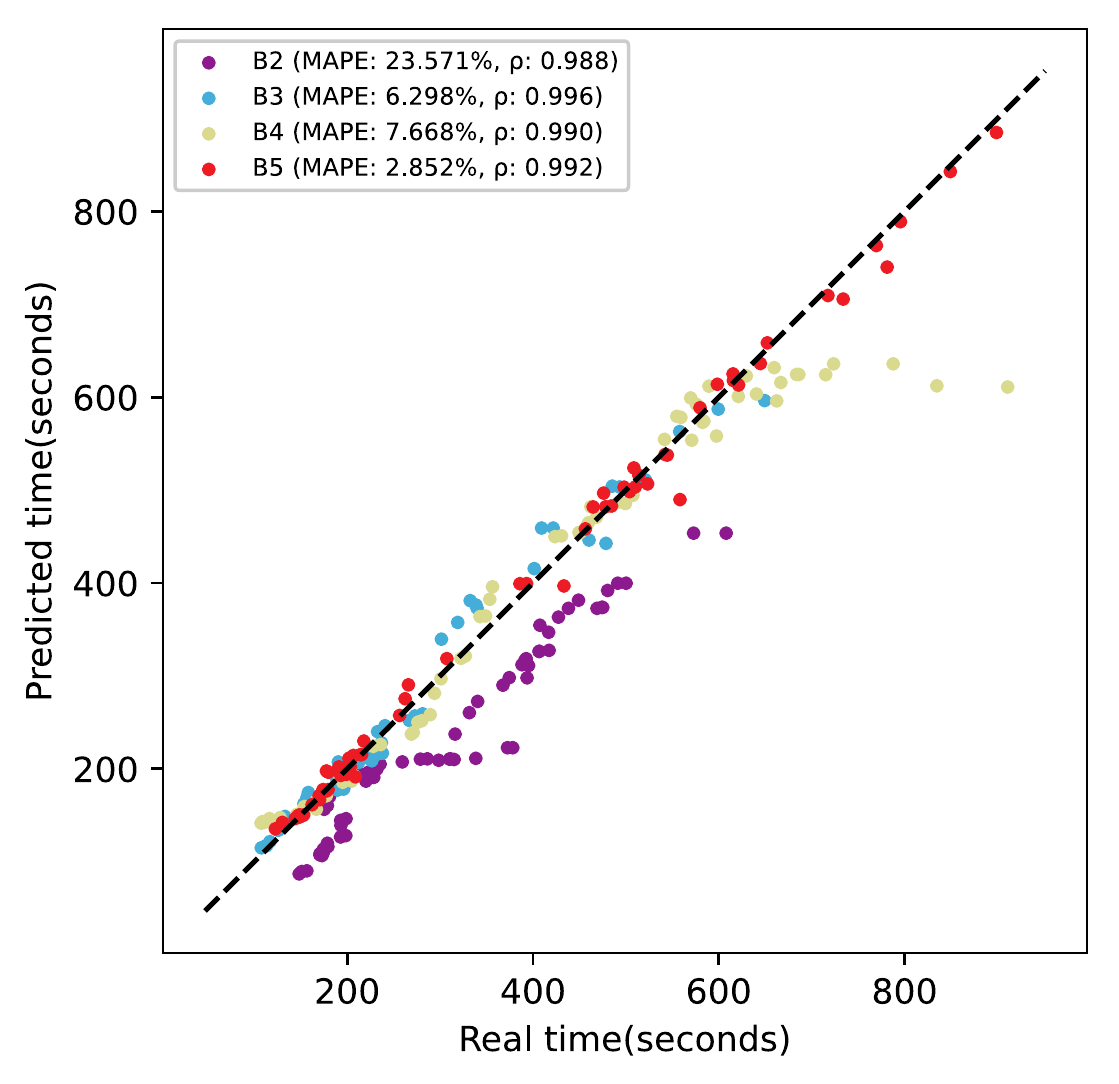}
	\caption[Accuracy predictor results]{Results of the time predictor on CIFAR10 dataset. The x axis indicates the training time required for training the cells, while the y axis indicates the training time estimated by the predictor.}
	\label{fig:time_results}
\end{figure}

From both the considered metrics perspectives, we notice CatBoost is outperforming all the other proposed methods.
In particular, we observe evident benefits which are visually observable in Fig.~\ref{fig:time_results} concerning the estimations qualities for the higher values of $b$.
This behavior is due to the increased amount of data, which CatBoost exploits to learn the correct time forecast for the cells explored during the search.

With CatBoost we have low MAPE, furthermore the Spearman coefficient is next to 1 in all steps, proving that the predictor is ranking the architectures correctly. Since ranking is the most desirable quality for accurately pruning in the Pareto front, we decided to use CatBoost as a time predictor in all the following experiments.

\section{Results}
This section presents the experiments conducted to evaluate POPNASv2 and compare it with PNAS.
In order to achieve a fair performance comparison, both the algorithms have been set up with the same search space.
Since there is no open-source version of PNAS search method, the tested version is derived by stripping the time predictor, Pareto front generation and exploration step from POPNASv2 method. The control step for pruning equivalent models is instead maintained. The hyperparameters common to both POPNASv2 and PNAS have the same values. The experiments have been carried out on a NVIDIA A100 GPU, using MIG 3g.20gb profile.

\subsection{Experiments setting}
The experiments conducted to validate the generalization capabilities of POPNASv2 efficiency improvements have been carried out on four different datasets for image classification, i.e., CIFAR10, CIFAR100~\cite{Krizhevsky09learningmultiple}, Fashion MNIST~\cite{xiao2017fashionmnist} and EuroSAT~\cite{helber2019eurosat}. These datasets differ in the number of classes, the number of channels, and the images dimensions, providing a robust benchmark for the generalization capabilities of the search strategy.

All the datasets share the same settings for preprocessing and data augmentation. 
In the preprocessing step, all input channels are normalized in [0, 1] range, and the samples are split into training-validation sets, respectively containing 90\% and 10\% of the total training samples. 
The default batch size is 128. We use random horizontal flip and random translation on both height and width for data augmentation, with a range of 0.125 as the actual input size.

The configuration of the search algorithm is the same for all these datasets. We train the architectures on the training set, keeping the validation to gather $ a $ and $ t $, used in the performance estimation strategy. The architectures are trained on $ E = 21 $ epochs, using AdamW~\cite{loshchilov2019decoupled} with cosine decay restart~\cite{loshchilov2016sgdr}, starting learning $ = 0.01 $, starting weight decay $ = 5e^{-4} $, $ T_{0} = 3 $ and $ T_{mul} = 2 $, for a total of 3 restart periods.
Swish~\cite{ramachandran2017searching} activation function is used instead of ReLU since it provided an accuracy boost to the trained architectures both in POPNASv2 and PNAS. 

Each training step after $ b = 1 $ trains at max $ K = 128 $ architectures. The maximum number of extra architectures selected in exploration step $ J $ is set to 16.
The neural networks produced starting from each cell are composed as a stack of 3 motifs, with $ N = 2 $, for a total of 8 stacked cells. This is the same structure also used in PNAS and NASNet.

The accuracy predictor is built as an ensemble of 5 LSTM using Attention. 
Each LSTM is trained for 30 epochs on $ \frac{4}{5} $ of the available data using Adam~\cite{kingma2017adam} optimizer, with $ lr = 4e^{-3}$ and L2 weight regularization with factor $ 1e^{-5} $. 
The embedding size used is composed of 10 units, the Conv1D filters are 16, and the cells used in each LSTM of the bidirectional are 48.

For each dataset, we used Catboost Regressor as time predictor, and tuned it via random search hyperparameters optimization.
The search space is defined as follows:
\begin{itemize}
    \item learning\_rate: uniform(0.02, 0.2)
    \item depth: randint(3, 7)
    \item l2\_leaf\_reg: uniform(0.1, 5)
    \item random\_strength: uniform(0.3, 3)
    \item bagging\_temperature: uniform(0.3, 3)
\end{itemize}

We trained each Catboost model for 2500 iterations in a 5-fold fashion, using early stopping with patience set to 50 iterations. The final model is retrained from scratch on the entire dataset, using the best hyperparameters configuration found.

\subsection{Predictors results}
\begin{table}[t!]
	\caption{The results of POPNASv2 accuracy predictor for each evaluated dataset.}
	\centering
	\resizebox{\linewidth}{0.9\height}{%
    	\begin{tabular}{|c||cccc||cccc|}
    		\hline
    		\multirow{2}{*}{Dataset} & \multicolumn{4}{c||}{MAPE(\%)} & \multicolumn{4}{c|}{Spearman($ \rho $)} \\
    		& b=2 & b=3 & b=4 & b=5 & b=2 & b=3 & b=4 & b=5 \\
    		\hline
    		CIFAR10 & 2.886 & 3.505 & 1.374 & 1.704 & 0.678 & 0.95 & 0.889 & 0.915 \\
    		CIFAR100 & 13.997 & 3.645 & 3.382 & 5.409 & 0.509 & 0.75 & 0.814 & 0.934 \\
    		Fashion MNIST & 2.296 & 0.891 & 1.067 & 0.861 & 0.798 & 0.875 & 0.84 & 0.928 \\
    		EuroSAT & 0.446 & 0.636 & 0.552 & 0.564 & 0.739 & 0.688 & 0.823 & 0.922 \\
    		\hline
    	\end{tabular}%
	}
	\label{table:accuracy_predictors_results}
\end{table}

\begin{table}[t!]
	\caption{The results of POPNASv2 time predictor for each evaluated dataset.}
	\centering
	\resizebox{\linewidth}{0.9\height}{%
    	\begin{tabular}{|c||cccc||cccc|}
    		\hline
    		\multirow{2}{*}{Dataset} & \multicolumn{4}{c||}{MAPE(\%)} & \multicolumn{4}{c|}{Spearman($ \rho $)} \\
    		& b=2 & b=3 & b=4 & b=5 & b=2 & b=3 & b=4 & b=5 \\
    		\hline
    		CIFAR10 & 23.571 & 6.298 & 7.668 & 2.852 & 0.988 & 0.996 & 0.99 & 0.992 \\
    		CIFAR100 & 21.661 & 9.202 & 6.881 & 3.709 & 0.964 & 0.975 & 0.951 & 0.989 \\
    		Fashion MNIST & 23.45 & 11.916 & 4.884 & 8.706 & 0.919 & 0.983 & 0.99 & 0.982 \\
    		EuroSAT & 25.721 & 9.031 & 9.913 & 5.937 & 0.963 & 0.974 & 0.987 & 0.97 \\
    		\hline
    	\end{tabular}%
	}
	\label{table:time_predictors_results}
\end{table}

The results of POPNASv2 predictors over the four datasets are summarized in Tab.~\ref{table:accuracy_predictors_results} and Tab.~\ref{table:time_predictors_results}. 
MAPE and Spearman values are computed on the unseen data, e.g., the columns $ b = 4 $ refer to the values forecasted starting from the information related to $b<4$.

In general, both predictors have less accurate results on $ b = 2 $ since the data available is fewer and the cells with $ b = 1 $ do not exhibit behaviors proper of multi-blocks cells.
In particular, cells constituted of multiple blocks could perform a concatenation operation followed by pointwise convolution at the end of the cell, used to join the outputs in case there are multiple unused block outputs. 
The presence of this extra pointwise convolution has a significant impact on the training time, leading to the error bias seen in the time predictor for $ b = 2 $. 
Moreover, some blocks could use other blocks as hidden layers to build a multi-level cell DAG, which is not possible in $ b = 1 $. 
These structural changes can alter the trend exhibited by time and accuracy metrics in single block cells, making $ b = 2 $ the most challenging step for predictors.
With the increase of $ b $, the average errors of the predictors tend to decrease, also improving the ranking quality.

\begin{figure}[t!]
	\centering
	\includegraphics[width=0.75\linewidth]{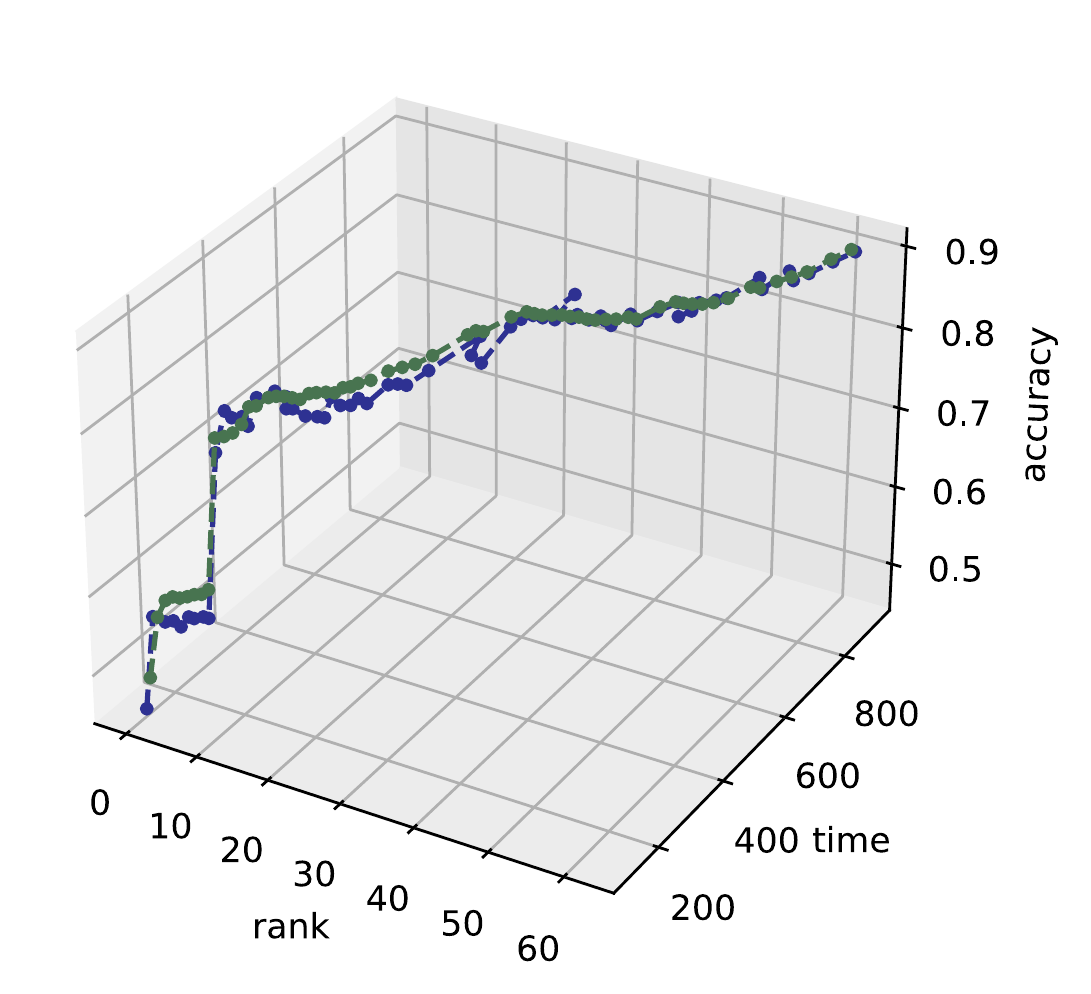}
	\caption[POPNASv2 CIFAR10 b=5 Pareto front]{The Pareto front trained for $ b = 5 $ in CIFAR10 dataset. The predicted values $ \hat{a}, \hat{t} $ are plotted in green, while the blue points represent the actual $ a, t $ retrieved after training. The rank values represent the ordering of these points in the Pareto front.  }
	\label{fig:pareto_plot_b5}
\end{figure}

The Spearman coefficients of both predictors tend to 1, so the predictors are ranking correctly the architectures, which is beneficial for POPNASv2 Pareto front methodology. 
From Fig.~\ref{fig:pareto_plot_b5}, which represents the Pareto front for $b=5$ over CIFAR10 dataset, it is possible to notice the quality of predicted performance with respect to the real ones.
We argue that, despite the low correlation between training time and accuracy achieved by a neural network, the features we selected for our predictors successfully allow POPNASv2 to build very accurate multi-objective rankings.

\subsection{POPNASv2 vs PNAS}
Tab.~\ref{table:search_results} shows a summary of the search results.
The results indicate that POPNASv2 is much more efficient than PNAS: with a 3.88x average time speed-up to complete the search, our algorithm can achieve the same average top-1 and top-5 accuracy performance concerning datasets differing in the number of classes, in the number of channels and in the dimension of the images. 
Contrarily to PNAS that always train $ K $ architectures in training steps with $ b > 1 $, POPNASv2 exploits the Pareto front pruning to drastically reduce the number of trained networks up to the 33.14\%.
Time optimization performed by POPNASv2 significantly lower the average training time of the considered networks, leading to a significant time speed-up also in the case both algorithms would train an equal amount of networks.

Surprisingly, for CIFAR100 and Fashion MNIST datasets POPNASv2 can also produce architectures achieving slightly better accuracy compared with the best ones produced by PNAS.
In general, the accuracy difference is at worse under one percentage point, confirming that our method can obtain remarkable gains from the accuracy-time trade-off.

\begin{figure*}[t!] 
	\centering
	\subfigure{\includegraphics[width=0.37\linewidth]{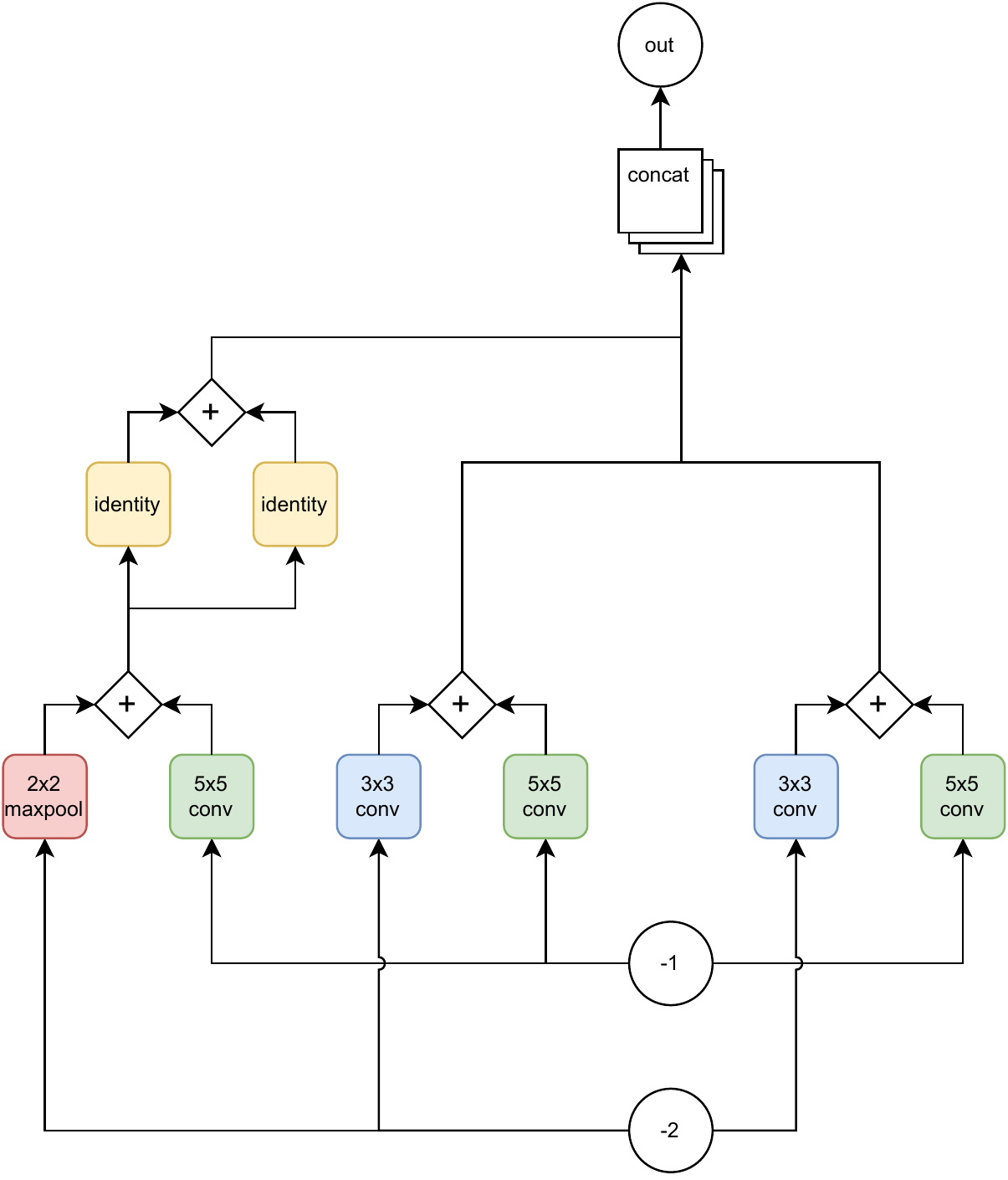}}
	\hfill
	\subfigure{\includegraphics[width=0.59\linewidth]{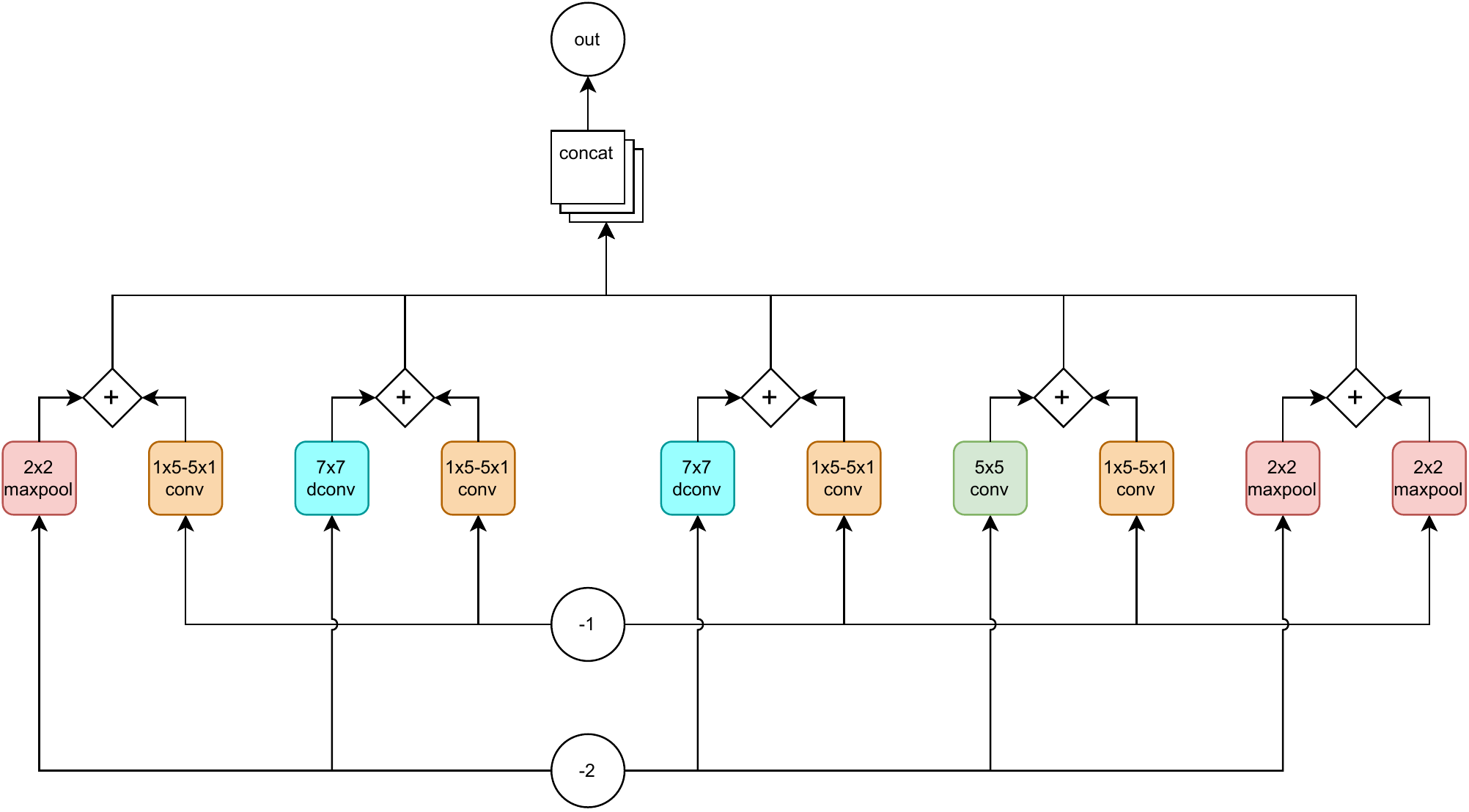}}
	\caption{Top1 cells found by POPNASv2 (on the left) and PNAS (on the right) over CIFAR10 dataset. The comparison shows how the chosen operators and their interconnections differ significantly between the two algorithms.}
	\label{fig:cells-cifar10}
\end{figure*}

The cell structures found by the two methods have some significant differences. Fig.~\ref{fig:cells-cifar10} provides a visual comparison among the ones found over CIFAR10 dataset.

POPNASv2 tend to organize cells into graphs composed of multiple levels, while PNAS cells are generally flat or almost flat. This behavior is imputable to the exploration step, which guarantees that the input values related to blocks are used in at least a small amount of networks, in each training step. The predictors can then adapt their results in following expansion steps, considering more these inputs values if they lead to good networks. PNAS does not guarantee the exploration of these input values, therefore the accuracy predictor tends to highly prefer inputs coming from other cells rather than from inner blocks.

\begin{table}[t!]
	\caption{The comparison between POPNASv2 and PNAS search performance over the evaluated datasets.}
	\centering
	\resizebox{\linewidth}{0.9\height}{%
    	\begin{tabular}{|cccccc|}
    		\hline
    		\multirow{2}{*}{Dataset} & \multirow{2}{*}{Method} & \multirow{2}{*}{\# Networks} & Top1 & Top5 & \multirow{2}{*}{Search Time} \\
    		& & & Accuracy & Accuracy & \\
    		\hline
    		\multirow{2}{*}{CIFAR10} & POPNASv2 & $ \mathbf{543} $ & 0.912 & 0.911 & $ \mathbf{49h 24m} $ \\
    		& PNAS & 814 & $ \mathbf{0.922} $ & $ \mathbf{0.92} $ & 176h 30m \\
    		\hline
    		\multirow{2}{*}{CIFAR100} & POPNASv2 & $ \mathbf{548} $ & $ \mathbf{0.685} $ & $ \mathbf{0.684} $ & $ \mathbf{53h 37m} $ \\
    		& PNAS & 814 & 0.68 & 0.679 & 161h 47m  \\
    		\hline
    		\multirow{2}{*}{Fashion MNIST} & POPNASv2 & $ \mathbf{537} $ & $ \mathbf{0.947} $ & $ \mathbf{0.946} $ & $ \mathbf{49h 50m} $ \\
    		& PNAS & 814 & 0.946 & 0.945 & 174h 53m \\
    		\hline
    		\multirow{2}{*}{EuroSAT} & POPNASv2 & $ \mathbf{549} $ & $ \mathbf{0.973} $ & 0.971 & $ \mathbf{69h 35m} $ \\
    		& PNAS & 814 & $ \mathbf{0.973} $ & $ \mathbf{0.973} $ & 350h 16m \\
    		\hline
    	\end{tabular}%
	}
	\label{table:search_results}
\end{table}

Another difference is in the choice of the operators. 
POPNASv2 tends to discard the operators that heavily impact the training time on the device hardware, except when they have the best performance among the entire operator set. 
Separable convolutions give a quite counter-intuitive example of this behavior. Even if they have fewer parameters and FLOPS than normal convolutions, they impacted a lot more on the training time. 
The inefficiency in parallelizing these operators could be the cause of the observed behavior.

Since PNAS is more targeted in finding top accuracy networks, it tends to choose complex and time-consuming operators. 
We further investigated the comparison between POPNASv2 and PNAS approaches by running an entire training session for each of their top-1 architecture until convergence.
In detail, we have retrained from scratch the best cells found by PNAS and POPNASv2 on the same dataset on which they were found, changing only the number of epochs $ E $ to 254 and reducing cosine decay restart $ T_{0} $ to 2. This experiment provides accurate results on the performance of the networks, using the same hyperparameter set used for the search method.

The results are summarized in Tab.~\ref{table:full_training_results}. 
Also in this scenario, POPNASv2 is able to fill the average accuracy GAP with respect to PNAS.
Concerning the top-1 networks training time, our approach achieved an average speed-up of 2.6x despite the slightly higher parameters number.
We argue that this behavior is caused by time optimization, which tends to prefer the operators which can be performed efficiently on the used hardware.
Concerning the number of blocks, we noticed that POPNASv2 tends to find cells composed by fewer blocks.

We can thus summarize the comparison, asserting that our approach can solve image classification problems by discovering simpler neural network architectures that achieve comparable performance with remarkable searching and training time speed-ups.

\begin{table}[t!]
	\caption{The comparison between POPNASv2 and PNAS top-1 networks performance over the evaluated datasets.}
	\centering
	\resizebox{\linewidth}{!}{%
    	\begin{tabular}{|cccccc|}
    		\hline
    		Dataset & Method & Params & B & Accuracy & Training Time \\
    		\hline
    		\multirow{2}{*}{CIFAR10} & POPNASv2 & 2.87M & 4 & 0.929 & $ \mathbf{1h 52m} $ \\
    		& PNAS & $ \mathbf{2.36M} $ & 5 & $ \mathbf{0.936} $ & 4h 6m \\
    		\hline
    		\multirow{2}{*}{CIFAR100} & POPNASv2 & $ \mathbf{2.27M} $ & 5 & $ \mathbf{0.718} $ & $ \mathbf{1h 55m} $ \\
    		& PNAS & 3.99M & 5 & 0.711 & 3h 19m \\
    		\hline
    		\multirow{2}{*}{Fashion MNIST} & POPNASv2 & 1.68M & 4 & $ \mathbf{0.951} $ & $ \mathbf{1h 56m} $ \\
    		& PNAS & $ \mathbf{1.41M} $ & 4 & 0.95 & 3h 4m \\
    		\hline
    		\multirow{2}{*}{EuroSAT} & POPNASv2 & 1.54M & 4 & $ \mathbf{0.979} $ & $ \mathbf{2h 47m} $ \\
    		& PNAS & $ \mathbf{1.47M} $ & 5 & $ \mathbf{0.979} $ & 11h 34m \\
    		\hline
    	\end{tabular}%
	}
	\label{table:full_training_results}
\end{table}

\section{Conclusion}
In this work, we introduced POPNASv2, a sequential model-based optimization search strategy solving the efficiency problem by building a time-accuracy Pareto front and exploiting its optimality properties. 
We used two surrogate models to estimate the training time and accuracy of the cells while searching for expansions, allowing to prune suboptimal results.
The adaptive greedy exploration step and the predictors features re-engineerization allowed our approach to find effective architectures in a more efficient fashion.
Comparing the results with PNAS, one of the most credited methods in the literature and baseline of our work, we obtained almost a 4x speed-up in the search time and achieved similar accuracy in the top networks.

\section*{Acknowledgments}
The European Commission has partially funded this work under the H2020 grant N. 101016577 AI-SPRINT: AI in Secure Privacy-pReserving computINg conTinuum.

\bibliographystyle{unsrt}  
\bibliography{bibliography}
\end{document}